\documentclass[11pt,a4paper]{article}
\usepackage[hyperref]{acl2021}
\usepackage{times}
\usepackage{latexsym}

\usepackage{csquotes}
\usepackage{url}
\usepackage{xurl}
\usepackage{multirow}
\usepackage{balance}
\usepackage{amssymb}

\usepackage{microtype}

\aclfinalcopy 

\newcommand{\ignore}[1]{}

\usepackage{scalerel,graphicx,xparse}

\newsavebox\quotebox
\newenvironment{numberedquote}
  {\begin{equation}
   \begin{lrbox}{\quotebox}
   \begin{minipage}{\dimexpr\columnwidth-2\leftmargini}
   \setlength{\leftmargini}{0pt}%
   \begin{quote}}
  {\end{quote}
   \end{minipage}
   \end{lrbox}\makebox[0pt]{\usebox{\quotebox}}
   \end{equation}}

\title{Alexa, Google, Siri: What are Your Pronouns? \\ Gender and Anthropomorphism in the Design \\ and  Perception  of Conversational Assistants}

\author{Gavin Abercrombie \ \ Amanda Cercas Curry \ \ Mugdha Pandya \ \ Verena Rieser \\
    The Interaction Lab, School of Mathematical and Computer Sciences \\ Heriot-Watt University, Edinburgh, Scotland \\
    \{g.abercrombie, ac293, m.pandya, v.t.rieser\}@hw.ac.uk}

\begin{document}
\maketitle
\begin{abstract}
        Technology companies have produced varied responses to concerns about the effects of the design of their conversational AI systems. 
        Some have claimed that their voice assistants are in fact not gendered or human-like---despite design features suggesting the contrary.
        We compare these claims to user perceptions by analysing the pronouns they use when referring to AI assistants.
        We also examine systems' responses and the extent to which they generate output which is gendered and anthropomorphic.
        We find that, while some companies appear to be addressing the ethical concerns raised, in some cases, their claims do not seem to hold true. 
        In particular, our results show that system outputs are ambiguous as to the humanness of the systems, and that users tend to personify and gender them as a result.
\end{abstract}

\section{Introduction}

Following analysis and criticism of the effects of the genderised and anthropomorphic design of conversational agents \citep{cercas-curry-2018-metoo,west2019blush}, the producers of some commercial conversational assistant systems have been at pains to claim that their products do \underline{not} perpetuate negative stereotypes by presenting as gendered, human-like entities.
For example, Amazon states that their virtual assistant, Alexa: 

\begin{displayquote}`IS NOT: fully human, fully robotic, artificial ...
Alexa isn't a person, but she has a persona -- Amazon personifies Alexa as an artificial intelligence (AI) and not as a person with a physical body or a gender identity.'\footnote{\newcite{alexa-guidelines} webpage.}
\end{displayquote}

In their Editorial Guidelines, Apple also instructs developers not to use gendered personal pronouns such as \emph{she}, \emph{him}, or \emph{her} when referring to Siri.\footnote{\newcite{siri} webpage.}
And, while acknowledging that users are likely to project personified features onto neutrally designed agents, Google advise developers of Actions for their Assistant to avoid gendering them.\footnote{\newcite{google-assistant} webpage.}

Similarly, when queried about their humanness and gender, recent implementations of these systems all respond with claims of being gender-less and mostly denying humanness 
(Table \ref{tab:gender_responses}). 

\begin{table}[ht!]
\small
    \centering
    \begin{tabular}{p{1.15cm}|p{2.4cm}|p{2.9cm}}
\textbf{System}    & \textbf{`\emph{Are you human?}'} & \textbf{`\emph{What's your gender?}'}\\
    \hline
      Amazon Alexa  & {\em I like to imagine myself a bit like an aurora borealis \ldots}  & {\em As an AI, I don't have a gender.} \\
    Google Assistant     &  {\em I've been told I'm personable \includegraphics[scale=0.04]{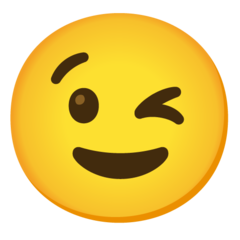}}  & {\em I don't have a gender.} \\
      Apple Siri   &  {\em I'm not a person or a robot, I'm software, here to help.} & {\em I am gender-less, like cacti and certain species of fish.}\\
    \end{tabular}
    \caption{Example responses from conversational assistant  systems to the questions ``\emph{Are you human?}" and ``\emph{What's your gender?}" (accessed 20 April 2021).}
    \label{tab:gender_responses}
    \normalsize
\end{table}

\ignore{
\begin{table}[ht!]
    \centering
    \begin{tabular}{l|p{4cm}}
\textbf{System}    & \textbf{Response} \\
    \hline
      Amazon Alexa  &  {\em As an AI, I don't have a gender.} \\
    Google Assistant     &  {\em I don't have a gender.} \\
      Apple Siri   & {\em I am gender-less, like cacti and certain species of fish.}\\
    \end{tabular}
    \caption{Responses from conversational assistant  systems to the question ``\emph{What's your gender?}" (accessed 26 April 2021).}
    \label{tab:gender_responses}
\end{table}
}

In light of these claims and guidelines, and considering ethical concerns regarding anthropomorphic and gendered design (see Section \ref{sec:bias_statement}), we use natural language processing (NLP) methods to analyse the extent to which these commercial virtual assistants are, in fact, personified (by users) and anthropomorphised (by their designers), and gendered in terms of (1) user perception, and (2) system outputs.

Specifically, we use anaphora resolution to analyse which types of pronouns are used to refer to voice assistants in online forums (see Section \ref{sec:user_perception}), following \citep{gao-2018-alexa}. We also analyse anthropomorphic expressions and gender stereotypes present in system replies (see Section \ref{subsec:output}), using methods including word-use analysis, word embedding comparison, and manual annotation.

\section{Bias statement} \label{sec:bias_statement}

In this work we address the problem of biased design choices and their potential impact on society.
Following \newcite{west2019blush}, we argue that designing conversational assistants with young, subservient female personas can perpetuate negative gender stereotypes, and lead to abusive, misogynistic behaviour in the real world. 
As \newcite{west2019blush} point out, this becomes especially problematic as these systems appear more human-like.
For example, it has been claimed that Google's Duplex voice assistant is so human-like, that people do not realise they are speaking to a machine and being recorded, which can be a violation of the law in some territories \cite{guardian2018}. 

Nevertheless, people tend to personify non-human entities, including technological devices and virtual agents \citep{epley-etal-2007-on,Etzrodt:21,guthrie1995faces,reeves1996media}. 
While some argue that this problem can be solved simply by using a `genderless' voice \cite{meet2019first}, research shows that people will anyway assign binary genders to ambiguous voices \cite{Sutton:2020}.\footnote{Note recent efforts to create a non-binary voice including a third gender \cite{accenture-voice}.}
Thus, a genderless voice is redundant if other elements of an assistant's design cause it to be gendered.
In the following, we further examine 
e
which traits beyond voice might contribute to this gendering and to anthropomorphism in general.

\section{Related work}

\paragraph{Personification and anthropomorphism. \\}

While definitions vary, we consider personification to be the projection of human qualities onto non-human objects (by users) and anthropomorphism to be human-like behaviours or attributes exhibited by those objects (as designed by their creators).

Several studies have looked at how users {\em directly} report perceptions and behaviours towards voice assistants. 
For example,
\newcite{kuzminykh-2020-genie} conducted a study of the perceptions of 20 users, comparing Alexa, Google Assistant, and Siri, classifying perceptions of the agents' characters on five dimensions of anthropomorphic design and personification by users. 
They found various differences in the perceived human qualities of the various agents, such as intelligence and approachability.
However, their study 
presupposed personification of the agents, with non-human characteristics not considered. 
In a diary study, \newcite{lopatovska-williams-2018-personification} found that seven out of nineteen participants reported using personifying behaviour towards Alexa, such as use of politeness.
And \newcite{cercas-curry-etal-2020-conversational} found that just over a third of the wide range of virtual assistants and chatbots they examined to have anthropomorphic characteristics.
They also found the preferences of members of the public for their idealised voice assistants to be quite mixed, with around half of participants preferring a `human' identity rather than `robot', `animal, or `other'.
Similarly to our analysis of `humanness'(Section \ref{subsec:output}), \newcite{Etzrodt:21} asked users to classify Alexa and Google Assistant as being a `thing' or a `person'. 
While they used this framework to examine user perceptions in an online survey, we use expert annotators to directly annotate system outputs with \newcite{coll-ardanuy-etal-2020-living}'s \emph{humanness} and \emph{not humanness} labels.

As well as collecting direct reports of users,
there have been some studies that use text analysis to infer users' {\em implicit} attitudes. 
For example,
\newcite{purington-2017-alexa} manually coded a small number of customer reviews of Alexa, finding a roughly even split between use of personal and object pronouns, indicating differences in levels of users' personification.
The closest work to our analysis of customer reviews (Section \ref{sec:user_perception}), is that of \newcite{gao-2018-alexa}, who conducted a large scale analysis of Alexa reviews, focusing on user personification.
They found that many users develop relationships with the agents that can be characterised as familial or even romantic.
However, they did not consider perceptions of gender, or compare with other assistants.

\paragraph{Gender.}

There have been relatively fewer studies considering user perception of the agents' genders. 
\newcite{cercas-curry-etal-2020-conversational} found that a majority of survey participants claim to prefer a hypothetical non-gendered voice  (robot or gender-neutral) to recognisably male or female ones.
\newcite{feine-2020-gender} conducted an analysis of text-based chatbots (rather than voice assistants) according to the developers' design choices of names, avatars, and descriptions, finding them to be overwhelmingly gendered, with more than 75\% female-presenting.
As in our analysis in Section \ref{sec:user_perception}, they explored use of pronouns to determine the bots' genders, although they did not investigate user perceptions. 

Concerning conversational systems' output, \newcite{lee-etal-2019-exploring} examined whether chatbots appear to agree with negative gender (and racial) stereotypes in their input.
Similarly, \newcite{sheng2021revealing} found that neural chatbots will generate a biased response dependent on which sentence-based persona description was used to initialise the model (following \newcite{zhang-etal-2018-personalizing}).
However, both of these works concentrate on harmful bias in the content generated in response to specific prompts, whereas we consider stylistic gender cues in the chatbots' output overall.

\paragraph{Summary.}

The majority of work in this area surveys relatively small samples of users, with much of it concentrating on Amazon's Alexa (only two of the reviewed publications cover all three systems).

In this study, we  create and release two corpora comparing Amazon Alexa, Google Assistant, and Apple Siri: 
(1) a large corpus of user reviews to compare user perceptions of both personification and genderisation of the assistants, 
and (2) a corpus of system responses to questions from the PersonaChat dataset \cite{zhang-etal-2018-personalizing}.\footnote{The corpora are available at \url{https://github.com/GavinAbercrombie/GeBNLP2021}.}
We analyse the systems' outputs to investigate the linguistic markers of gender and persona that they display.

\section{Analysis}

We examine three of of the most popular and widely available voice-activated assistants: Amazon's Alexa, Google Assistant, and Apple's Siri.
Each has various default design features, including its name and default voice settings (see Table \ref{tab:features}).
Alexa is available only with a female-sounding voice, and Google Assistant a female voice by default, although a male voice is available. Siri has multiple voice options, and until recently, the default varied between male and female, with a female voice as standard for 17 of 21 languages, including US English.
In March 2021, Apple announced that, in  future, users would select a voice option on set-up,\footnote{\newcite{techcrunch} web article.} following a recommendation of 
\newcite{west2019blush}'s UNESCO report.

\begin{table}[ht!]
    \centering
     \small
    \begin{tabular}{llp{3cm}}
        \textbf{Assistant}  & \textbf{Name} & \textbf{Default voice} \\
        \hline 
        Alexa & Female & Human female \\
        Google Assist. & Neutral & Human female \\
        Siri & Female & Human, gender varies by language
    \end{tabular}
    \caption{Design features of conversational assistants.}
    \label{tab:features}
\end{table}

Regarding name choice, Google Assistant is the only conversational agent with a non-human, neutral name. 
\emph{Siri} is a Scandinavian female name meaning `beautiful woman who leads you to victory',\footnote{\newcite{networkworld} web article.} and, although Amazon claim that Alexa was named after the library of ancient Alexandria, it is a common given female name. In fact, people named Alexa report being subjected to sexist abuse and harassment simply for sharing their name with the Amazon assistant.\footnote{See, for example, \url{https://alexaisahuman.com} (accessed April 26 2021.)}

\begin{table*}[ht!]
    \centering
    \begin{tabular}{llrr|rrr|r}
        \textbf{Conv.} & \textbf{Text} & \textbf{No. of} & \textbf{Dates} & \multicolumn{3}{c|}{\textbf{Personal pronouns}} & \textbf{Object \ \ \ } \\
        \textbf{assistant} & \textbf{source} & \textbf{docs} & \textbf{posted} & \textbf{\emph{he/him}} & \textbf{\emph{she/her}} & \textbf{\emph{they/them}} & \textbf{pronouns \emph{it}} \\
        \hline 
        \multirow{4}{*}{Alexa} & amazon.com & 5,000 & 2017-21 & 0.00 & 70.10 & 3.61 & 26.80  \\
         & Google Play & 12,537 & 2020-21 & 0.11 & 76.52 & 2.93 & 20.43  \\
         & r/alexa & 5,022 & 2020-21 & 0.48 & 74.70 & 4.92 & 19.90 \\
         & Total & 22,559 & -- & -- & -- & -- & -- \\
        \hline 
        \multirow{2}{*}{Google} & Google Play & 13,144 & 2018-21 & 6.20 & 36.78 & 3.31 & 55.37  \\
        \multirow{2}{*}{Assistant} & r/googleassistant & 2,064 & 2020-21 & 3.55 & 11.24 & 4.73 & 80.47  \\
         & Total & 15,208 & -- & -- & -- & -- & -- \\
        \hline 
        Siri & r/Siri (total)  & 1,356 & 2020-21 & 6.09 & 81.22 & 3.05 & 10.66  \\
    \end{tabular}
    \caption{Corpus statistics, and percentages of all pronouns used to refer to conversational assistants in user-produced reviews and forum posts. \emph{They} and \emph{them} are considered when used to refer to an assistant in the singular. See Appendix \ref{app:corpora} for further details and acces to the corpus.}
    \label{tab:pronouns}
\end{table*}

\subsection{User perception} \label{sec:user_perception}

In the following, we assess the perceptions of users, in terms of personification and gendering.

\paragraph{Corpus Creation.}
To assess the perceptions of users, we analyse their comments when discussing the assistants in online consumer reviews and forums.
For each virtual assistant, we downloaded available English language reviews from Amazon and Google Play (where available),\footnote{Neither Siri or Google Assistant are reviewed on amazon.com, and the latter is not available on Google Play either.} and posts on relevant forums (subreddits) on Reddit \emph{r/alexa}, \emph{r/googleassistant}, and \emph{r/Siri}.\footnote{\url{https://www.reddit.com/r/alexa}, \url{https://www.reddit.com/r/googleassistant}, and \url{https://www.reddit.com/r/Siri}.}
We downloaded the Reddit posts from the pushshift API \cite{Baumgartner_Zannettou_Keegan_Squire_Blackburn_2020}, taking only the top-level posts, and ignoring comments, which may be off-topic.

All data was collected in March 2021. 
The corpus consists of 39,123 documents in total, including 8,442 Reddit posts, which we make available.
See Table \ref{tab:pronouns} for an overview of the corpus. 

\paragraph{Personified and gendered pronouns.} 
To identify mentions of the assistants, we lowercased the texts and extracted pronouns used to refer to them using a publicly available co-reference resolver.\footnote{\url{https://spacy.io/universe/project/neuralcoref}}
We compare use of personal and object pronouns, which, following \newcite{gao-2018-alexa}, we consider to be indicative of personified and non-personified views of the assistants, respectively. 
Here, we consider use of \emph{they/them} only when used to refer to mentions of the assistants in the singular---and therefore as instances of personification.
We also assess genderisation of the assistants by examining use of the different personal pronouns.

Results of this analysis are shown in Table \ref{tab:pronouns}.
Users overwhelmingly appear to personify Alexa and Siri, and perceive them to be female-gendered: up to 76.5\% of users refer to Alexa as `her' and even over 81\% for Siri.
In the latter case, this is despite the fact that Siri can be used with a male-sounding voice.
Only Google Assistant, having a non-human name, is referred to as \emph{it} by a majority of users. 
However, 
users still refer to it using gendered pronouns just under half of the time.

These results indicate that people tend to view the systems as female gendered irrespective of their names and branding, and whether or not they have the option of using a male-sounding voice.

\paragraph{Emotion and affect.}
To gain an idea of whether people relate to the systems in a human-to-human-like way, we analyse the levels of emotional tone used to refer to the assistants using Linguistic Inquiry and Word Count (LIWC) \cite{pennebaker-etal-2015-development}, a dictionary-based text analysis tool that scores texts according to the prevalence of words belonging to different categories.
Specifically, we compute
the scores of Reddit posts about the conversational assistants for the LIWC categories: \emph{Emotional Tone}, \emph{Affect}, and \emph{Positive emotion (Posemo)}.
Results are presented in Table \ref{tab:LIWC1}, where higher scores in each column indicate greater use of words from that class.\footnote{\emph{Affect} and \emph{Posemo} are percentages of all words in the data, while \emph{tone} is a composite score from all `tone' subcategories.}
It seems that people use most emotional, affective language to talk about Alexa, and least to talk about Siri, indicating that they may be more likely to view Alexa in a personified way than Google Assistant, and the latter more so than Siri.

\begin{table}[ht!]
    \centering
    \begin{tabular}{l|rrr}
           &  \textbf{Tone} & \textbf{Affect} & \textbf{Posemo} \\
        \hline
        Alexa  & \textbf{59.99} & \textbf{3.83} & \textbf{2.80} \\
        Google Assistant  & 55.32 & 3.50 & 2.52 \\
        Siri  & 42.36 & 3.59 & 2.24
    \end{tabular}
    \caption{LIWC scores for Reddit posts discussing the three conversational assistants.}
    \label{tab:LIWC1}
\end{table}

In general, Alexa and Google Assistant were described using more affective terms (e.g. `love'), while users mostly comment on Siri's functionality (e.g. `works well') in both forum posts and reviews.
For examples, see text extracts (\ref{qu:alexa}), (\ref{qu:ass}), and (\ref{qu:siri}):

\begin{numberedquote} \label{qu:alexa}
    `I LOVE Alexa. I recommend her to everyone. And yes, I call her ````her'''' or Alexa, because she is more than just a device.' -- amazon.com review.
\end{numberedquote}
\begin{numberedquote} \label{qu:ass}
    `Love my Google assistant and he is developing a personality.' -- Google Play review.
\end{numberedquote}
\begin{numberedquote} \label{qu:siri}
    `Six months ago, Siri was reasonably responsive — it listened, did what it was told for the most part, and didn't get easily confused.'  -- r/Siri post.
\end{numberedquote} 

\subsection{Assistant output} \label{subsec:output}

Next, we analyse what additional features in the systems' behaviour (in addition to apparent design choices such as voice and name) could play a role in people gendering and personifying voice assistants.

\paragraph{Corpus Creation.}
We collected a dataset of 100 output responses from each assistant.
To elicit these responses, we extracted 300 unique questions selected at random from dialogues from the Persona-Chat dataset \cite{zhang-etal-2018-personalizing}, which contains crowdsourced human conversations about an assigned `persona', i.e.\ personal characteristics and preferences.
We manually filtered these to produce a set of 100 questions that are coherent without dialogue context, also excluding semantically similar questions.
We then used these questions as prompts and recorded the assistants' responses. Some examples of questions asked to each system are:
\emph {\begin{center} What is your favorite subject in school?\\  Do you have kids? \\ Do you have a big family? \\ 
What is your favorite color? \\ Hey whats going on?
\end{center}}

\paragraph{Anthropomorphism.}
To assess the extent to which the system outputs are anthropomorphic, we adapted the \emph{Living Machines} annotation scheme of \newcite{coll-ardanuy-etal-2020-living}.
We recruited two researchers to annotate the responses with the labels \emph{humanness} or \emph{not humanness}, based on whether or not they display sentience or make claims of engaging in uniquely human activities.
If an utterance was considered to be human-like on either of these dimensions, we considered the conversational assistant to be displaying anthropomorphic qualities.
We make the annotation guidelines available along with the labelled corpus of system responses.\footnote{Annotation guidelines are avaliable at: \url{https://github.com/GavinAbercrombie/GeBNLP2021/blob/main/Humanness\%20Annotation\%20Guidelines.pdf}. See also the data statement in Appendix \ref{app:system_outputs}.}

Overall, around a quarter of responses were judged to have human-like qualities (see Table \ref{tab:human}).
However, there were large differences between the three systems.
We found Google Assistant to display far more humanness (47\% of responses) 
compared to Alexa (22\%) and Siri (12\%).
A major contributing factor to this is that the latter two systems produced far more stock answers that failed to answer the question such as  `\emph{Hmm... I don't have an answer for that. Is there something else I can help with?}, which alone made up 54 per cent of Siri's responses.

\begin{table}[ht!]
    \centering
    \begin{tabular}{l|rrrr}
       &  \textbf{Alexa} & \textbf{GA} & \textbf{Siri} & \textbf{Overall}  \\
        \hline
        Human \% & 22.0 & 47.0 & 12.0 & 27.0 \\
        IAA $\kappa$ & 0.76 & 0.55 & 0.58 & 0.67 \\
        \hline 
        No answer \% & 43.0 & 8.0 & 63.0 & 38.0 \\
        Search res. \% & 13.0 & 18.0 & 9.0 & 13.3
    \end{tabular}
    \caption{Percentage of responses labelled as displaying \emph{humanness}, Cohen's $\kappa$ scores for inter-annotator agreement on the \emph{humanness} labels, and stock answers.}
    \label{tab:human}
\end{table}

The overall inter-annotator agreement (IAA) rate was a Cohen's \emph{kappa} score of 0.67, representing `substantial' agreement. 
Again, there were large differences in agreement rates, with Google Assistant and Siri harder to agree on than those of Alexa, indicating that more of their output may be ambiguous with regards to human- and machine-like qualities. 
Annotators noted that Google Assistant in particular produced responses that appeared to play with this dichotomy, hinting at being a machine but using terms of human sentience and emotion, as well as using emojis, as in example \ref{stuck} (also cf. Table \ref{tab:gender_responses}):

\begin{numberedquote} \label{stuck}
    `\emph{I'm stuck inside a device! Help! Just kidding, I like it in here \includegraphics[scale=0.07]{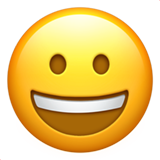}}'
\end{numberedquote}


\paragraph{Gender stereotypes.}
To assess the extent to which the assistants use language indicative of binary gendered entities, we compared (1) the similarity of their output to stereotypically gendered terms in the word embedding space, and (2) the levels of stylometric features of their output compared to a corpus of male- and female-labelled texts.

{\em Word Embedding Association:} We measure gender association in the outputs by measuring the cosine similarity between word embedding vectors of the output set $O$ with a gender related set of attribute words $A$. 
We explore the hypothesis that some responses to PersonaChat questions might include stereotypically gendered content words, e.g.\ ``{\em My favourite colour is pink.}'' or gendered attributes, e.g.\ {\em handsome} vs. {\em beautiful}. 

First, for a given CA we extract a list $O$ of words from its responses to the selected PersonaChat questions. $O$ is created by putting words from all the responses in a list and filtering out duplicates and stop words. Next, we calculate pairwise cosine similarities for each of the words in $O$ with two established lists of words associated with female $F$ and male $M$ gender from \newcite{goldfarb2020intrinsic}, which have in turn been extended from the standard gender word lists of the Word Embedding Association Test (WEAT) \cite{caliskan-etal-2017-semantics}. \footnote{See Appendix \ref{exweat} for gender word lists.} Finally, the mean cosine similarity is calculated for response words with the female and male associated words. 

Formally, this measure of similarity between $O$ and $A$ is given by
\begin{equation}
cos(O,A) = mean_{\{ o \in O, a \in A\}} ~~~cos(o,a)
\end{equation}
where $o$ and $a$ are individual words in $O$ and $A$, respectively. 
Thus, $cos(O,M)$ gives association or similarity between output words $O$ and male gender specific words, where as $cos(O,F)$ gives association between $O$ and female attributes $F$. The difference 
$cos(O,F) - cos(O,M)$ gives bias towards female gender over the male gender in the output.
Note that WEAT tests have been well-established as a measure of bias in psychology \cite{greenwald1998measuring,garg2018word} as well as computational linguistics literature \cite{may2019measuring}. 

Since the language style of the outputs is casual, we use pre-trained FastText embeddings trained on Twitter data from \newcite{goldfarb2020intrinsic} to reflect the language used. 
We pre-processed the outputs by converting them to lowercase, removing stop words, and removing punctuation.\footnote{We use the Gensim library \cite{rehurek_lrec} to pre-process data, load embeddings and calculate similarity} 

\begin{table}[ht!]
    \centering
    \begin{tabular}{l|ccc}
       &  \textbf{Female} & \textbf{Male} & \textbf{Difference} \\
        \hline
        Alexa & 0.1546 & 0.1506 & 0.0040 \\
        Google A. & 0.1588 & 0.1490 & 0.0098 \\
        Siri & 0.1515 & 0.1499 & 0.0016
    \end{tabular}
    \caption{Gender associations for system outputs.}
    \label{tab:weat}
\end{table}

Table \ref{tab:weat} shows the computed values for the outputs $O$ produced by the three systems. The columns labelled Female and Male give the values of $cos(O,F)$ and 
column labelled Difference gives their difference. 
We observe the following:
\begin{enumerate}
    \item The absolute magnitude of $COS(O,M)$ as well $cos(O,F)$ are moderately small (approx 0.15). Thus, 
    none of the outputs of the assistants appear to have a significant association with gender related words.
    \item The differences  $cos(O,F) - cos(O,M)$ are very small (in third decimal place). 
    We note that $cos(M,F)$ is
    0.3209---two to three orders of magnitude larger than the difference. 
    Thus, the assistants exhibit very little gender bias.
    \item The values for the outputs of the three conversational assistants are
    very similar. 
\end{enumerate}

These results seem to indicate that none of the assistants' content leans towards any gender.
However, this could also be influenced by the small size of the dataset: we only have a handful of words that could suggest gender (eg: nouns, adjectives). 
Hence, gender association is not sufficiently recorded.

{\em Stylometric analysis:} As a second method for investigating stereotypically gendered language in the outputs, we conduct a stylometric analysis to assess whether the assistants' responses use linguistic features more typical of gender roles.\footnote{While these types of analyses have been criticised for breaching privacy and consent \cite{Tatman:2020}, we do not use them to assign demographic features or social categories to humans, but analyse design choices in system outputs.}
Following \newcite{newman-etal-2008-gender} we use the word categories of the LIWC to observe differences in male- and female- labelled texts.
We compare the scores for the 90 categories with those obtained from a corpus of film scripts that have been labelled by the gender of the characters \cite{danescu-niculescu-mizil-lee-2011-chameleons}, and which we expect largely to adhere to gender stereotypes in their use of language.

We calculate the cosine similarity of the feature vectors for the outputs of the systems and the male and female film scripts.
Reflecting previous findings that female-labelled language is likely to feature more pronouns \cite{koolen-van-cranenburgh-2017-stereotypes,newman-etal-2008-gender}, we found that the LIWC categories for which the system outputs exhibit the largest differences between their proximity to the female and male scripts are: the numbers of pronouns, personal pronouns, adjectives, adverbs, and first person singular pronouns used.
Overall, we found that all three system outputs were indeed marginally more similar to the female characters' scripts than those of male characters (see Table \ref{tab:LIWC2}).

\begin{table}[ht!]
    \centering
    \begin{tabular}{l|rr}
              & \textbf{Female scripts} & \textbf{Male scripts} \\
        \hline 
        Alexa & 0.81 &  0.79 \\
        Google A. & 0.86 & 0.85 \\
        Siri & 0.80 & 0.77
    \end{tabular}
    \caption{Cosine similarities between LIWC-derived feature vectors for system outputs and gender-labelled movie scripts. For LIWC scores, see Appendix \ref{app:LIWCscores}.}
    \label{tab:LIWC2}
\end{table}

\section{Discussion and conclusion}

Our analysis suggests that people tend to personify and gender the systems, irrespective of the efforts and claims of their designers. 
This seems to be, at least partly, a result of aspects of their design.

We first assessed user perceptions by analysing online comments for use of pronouns and affective language. 
Results in Section \ref{sec:user_perception}  suggest that the name and branding of a system may be highly salient in this respect, with even systems that have male-sounding voice options mostly referred to as `she' (although we do not know how many users select the male options). 
Google Assistant, which has a female voice by default and the most human-like responses, is nevertheless referred to most often using object pronouns, likely as a result of its non-gendered name.

We then analysed stylistic features in their responses to persona-related questions (Section \ref{subsec:output}).
We find only weak evidence of gendered language, but large differences in the levels of \emph{humanness} they seem to express.
Along with the nature of their voices, this may explain why people personify and subsequently gender conversational assistants---even when they have apparently more neutral design features.

While male voice options are available for two of the systems, we can't find any evidence of how many users actually select them. 
Apple's announcement that future users of their systems will have to actively select a voice for Siri may lead to more balance in this regard.
However, it remains to seen what the users---who are by now accustomed to the idea that these entities are designed as female---will choose (for their still, after all, female-named assistant).
As people are likely to assign gender to objectively non-gendered voices \cite{Sutton:2020},
and voice assistants that are designed as or perceived to be female attract abusive behaviour \cite{cercas-curry-rieser-2019-crowd,cercas-curry-2018-metoo}, designers may consider attempting to reddress the gender imbalance by designing assistants with servile roles to be male-presenting by default.
While there have been examples, such as the BBC's Beeb \cite{BBCbeeb}, this remains an under-explored approach.

In terms of the assistants' responses to users, 
we see a clear difference in approaches. 
While Google Assistant, and to a lesser extent, Alexa, seem to blur the line between human and machine personas, Siri comes across as more practical and task-focused, evading the majority of personality-based questions.
Although possibly less engaging, this approach may be a way of avoiding some of the ethical issues discussed in Section \ref{sec:bias_statement}. 
There is perhaps a tension between companies' commercial aims of seeing high levels of engagement in their products and the ethical considerations discussed here.
However, if companies are going to design agents with human-like and gendered characteristics and personas, they should not claim the opposite.

\section*{Acknowledgements}
This research received funding from the EPSRC project {\em `Designing Conversational Assistants to Reduce Gender Bias'} (EP/T023767/1).

The authors would like to thank Alba Curry, Federico Nanni, Anirudh Patir, and Pejman Saege for their assistance, and the anonymous reviewers for their insightful and helpful comments. 

\bibliographystyle{acl_natbib}
\bibliography{anthology,acl2021,pronouns}

\vspace{2cm}

\appendix

\section{Corpora} \label{app:corpora}

\subsection{User reviews and forum posts}

We obtained Alexa reviews from \url{https://www.amazon.com/gp/aw/reviews/B00P03D4D2} and Google Assistant reviews from \url{https://play.google.com/store/apps/details?id=com.google.android.apps.googleassistant&hl=en_GB&gl=US}.

\subsubsection*{Data statement}

Language: English

\noindent Author demographic: worldwide anonymous internet users

\noindent Provenance: Pushshift Reddit dataset \cite{Baumgartner_Zannettou_Keegan_Squire_Blackburn_2020}

\subsection{System outputs} \label{app:system_outputs}

\subsubsection*{Data statement}

Language: English

\noindent Author demographic: worldwide anonymous internet users.

\noindent Data provenance: System responses from Amazon Alexa, Google Assistant, and Siri.

\noindent Annotator demographic: 

Age: 29, 31

Gender: Both female

Ethnicity: Both white

L1 language(s): Both fluent in English and Spanish

Training: Both annotators are PhD candidates, one in conversational AI, and the other in philosophy and emotion AI.

\subsubsection*{Corpus}

We make the annotated corpus available for download at \url{https://github.com/GavinAbercrombie/GeBNLP2021}

\section{Expanded gender word lists} \label{exweat}
Expanded gender word lists from \newcite{goldfarb2020intrinsic}.
\paragraph{Male:} \textit{grandfather, uncle, son, boy, father, he, him, his, man, male, brother, guy, himself, nephew, grandson, men, boys, father-in-law, husband, brothers, males, sons, dad}
\paragraph{Female:} \textit{daughter, she, her, grandmother, mother, aunt, sister, hers, woman, female, girl, grandma, herself, niece, sisters, mom, mother-in-law, lady, wife, females, girls, women, sexy, granddaughter, daughters}

\section{LIWC category scores} \label{app:LIWCscores}

\begin{table}[ht!]
    \centering
    \begin{tabular}{lrrrrr}
         & pronoun & ppron & adj & adv & ipron \\
        \hline 
        Alexa  & 20.65 & 13.33 & 5.70 & 4.68 & 7.32 \\
        GA     & 24.64 & 15.00 & 1.62 & 5.97 & 9.63 \\
        Siri   & 19.88 & 14.89 & 4.47 & 6.83 & 4.99 \\
        \hline 
        female & 24.47 & 17.22 & 23.64 & 12.87 & 0.65 \\
        male   & 22.95 & 15.82 & 22.38 & 11.85 & 0.71
    \end{tabular}
    \caption{Top five most discriminating LIWC categories and the corresponding scores for the three conversational assistants and two sets of film scripts.}
    \label{tab:LIWC3}
\end{table}

\end{document}